\documentclass[letterpaper, 10 pt, conference]{ieeeconf}  

\IEEEoverridecommandlockouts                              

\usepackage{amssymb}  
\usepackage{graphics} 
\usepackage{graphicx}
\usepackage[tight,footnotesize]{subfigure}
\long\def\invis#1{}
\usepackage{cite}
\usepackage{url}
\usepackage{hyphenat}
\usepackage{amsmath,systeme}
\usepackage{xcolor}
\usepackage{siunitx}

\newcommand\fig[1]{Figure~\ref{#1}}

\newcommand\etal{\textit{et al.\ }}

\usepackage{algpseudocode}
\usepackage{algorithm}

\def\floatsep{4.0pt}
\def\textfloatsep{0.0pt}
\def\dblfloatsep{4.0pt}
\def\dbltextfloatsep{4.0pt}

\usepackage{fp}
\FPset{\pb}{0}
\newcommand{\pagebudget}[1]{}
\newcommand{\showtotalpagebudget}{}

\title{\LARGE \bf Navigation in the Presence of Obstacles\\for an Agile Autonomous Underwater Vehicle 
}

\author{Marios Xanthidis, Nare Karapetyan, Hunter Damron, Sharmin Rahman, James Johnson,\\Allison O'Connell, Jason M. O'Kane, and Ioannis Rekleitis 
\thanks{A. O'Connell is with the Computer Science Department, Vassar College, Poughkeepsie, NY.  The  remaining authors are with the Computer Science and Engineering Department, University of South Carolina, Columbia, SC, USA
        {\tt\small [mariosx, nare, hdamron, 
srahman, jvj1]@email.sc.edu, aoconnell@vassar.edu, [jokane,yiannisr]@cse.sc.edu}}%
\thanks{This work was made possible through the generous support of  National Science Foundation grants (NSF 1513203, 1526862, 1637876, 1849291).}
}
\begin{document}

\maketitle
\thispagestyle{empty}
\pagestyle{empty}

\begin{abstract}
Navigation underwater traditionally is done by keeping a safe distance from obstacles, resulting in ``fly\hyp overs'' of the area of interest. Movement of an autonomous underwater vehicle (AUV) through a cluttered space, such as a shipwreck or a decorated cave, is an extremely challenging problem that has not been addressed in the past. This paper proposes a novel navigation framework utilizing an enhanced version of \emph{Trajopt} for fast 3D path-optimization planning for AUVs. A sampling-based correction procedure ensures that the planning is not constrained by local minima, enabling navigation through narrow spaces. Two different modalities are proposed: planning with a known map results in efficient trajectories through cluttered spaces; operating in an unknown environment utilizes the point cloud from the visual features detected to navigate efficiently while avoiding the detected obstacles. The proposed approach is rigorously tested, both on simulation and in\hyp pool experiments, proven to be fast enough to enable safe real-time 3D autonomous navigation for an  AUV.

\end{abstract}
\section{Introduction}\pagebudget{1}


This paper addresses the problem of trajectory planning for underwater structure inspection and mapping. Underwater structure mapping is an important capability applicable to multiple domains: marine archaeology, infrastructure maintenance, resource utilization, security, and environmental monitoring. While the proposed approach is not limited to the underwater domain, this work addresses the challenging conditions encountered underwater, with a special focus on shipwreck mapping. Underwater mapping has traditionally focused on acoustic sensors~\cite{gary20083d,johannsson2010imaging,teixeira2018multibeam,przeslawski2018comparative}.  However, most inspections require visual input~\cite{abadie2018georeferenced,drap2012underwater,figueira2015accuracy}. The state\hyp of\hyp the\hyp art in autonomous operations is to observe the target structure from far enough to avoid navigation in cluttered spaces, while remotely controlled operations present entanglement hazards. As a result, most inspections suffer from gaps due to occlusions, and low\hyp resolution due to the water effects on the camera sensor. The  presented  work enables autonomous operations underwater close to the structure to be inspected; before this was only partially possible with teleoperation.

Maps of underwater structures such as wrecks, caves, dams, and docks are often available either through acoustic sensing~\cite{leonard2012directed} or via photogrammetry~\cite{menna2018photogrammetric,agrafiotis2018underwater}. In this paper we present an augmentation of  Trajopt~\cite{schulman2013finding}, a popular path\hyp optimization open source package for (mobile) manipulators, to facilitate 3D trajectory planning  of an AUV utilizing either a known map or an online constructed local map. The proposed method is realized on an Aqua2 vehicle~\cite{Rekleitis2005d}. The Trajopt planner is augmented with a sampling\hyp based correction scheme that resolves the local minima problem. Furthermore, different map representations were tested including geometric primitives and point\hyp cloud implementations.  

\begin{figure}
\begin{center}
    \includegraphics[width=0.9\columnwidth]{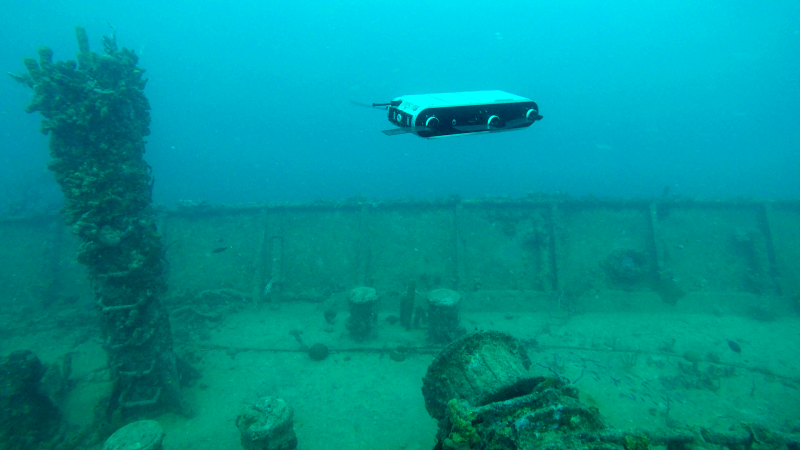}
    \caption{Aqua2 AUV navigating over the Stavronikita shipwreck, Barbados.}
    \label{fig:beauty}
\end{center}
\end{figure}

Fast motion planners, such as KPIECE~\cite{csucan2009kinodynamic}, use physics\hyp based simulators to plan with kinodynamic constraints and can solve the planning problem for systems with non\hyp trivial dynamics. While previous work~\cite{georgiades2009simulation,giguere2006characterization,plamondon2008trajectory} provided an analysis of the dynamics of the Aqua vehicle, the model is still inaccurate and moreover, the on\hyp board computing resources do not allow for utilizing a physics\hyp based simulator online.

The proposed method was rigorously tested in simulation and in the pool. Utilizing the Gazebo simulator~\cite{koenig2004design} with an underwater extension that emulates the kinematic behaviour of the Aqua2 vehicle, the tested environments demonstrated changes in depth, and attitude in three dimensions for realizing the produced trajectories. Furthermore, tests at an indoor diving pool, with various obstacle setups, verified the validity of the proposed approach.

The proposed approach provides contributions in two conceptual areas: path planning of mobile robots in 3D and underwater navigation in cluttered environments. More specifically, we augmented the Trajopt package enabling operations of an autonomous mobile robot in three dimensions. We introduced a fast warm\hyp starting method to avoid local minima issues using an RRT\hyp based approach. Furthermore, we demonstrated the use of Trajopt for online planning based on convex decomposition of unordered point clouds. The proposed framework enabled underwater operations in cluttered spaces. In particular, a light-weight geometric navigation framework utilizing the full 3D motion capabilities of the Aqua2 AUV, enabled operations with a known map, or in unknown areas of cluttered underwater environments. Autonomous operations of real\hyp time planning and replanning in conjunction with visual inertial state estimation in environments of substantial complexity, even without utilizing a known hydro-dynamics model.

\section{Related Work}\pagebudget{0.5}
\label{ch:litReview}
In environments with complex dynamics, such as underwater,  one of the main challenges is to generate safe paths. Several methods have been explored to correct the deviations caused by inertia and currents, including the FM* planning system~\cite{petres2007path}. Other methods rely on observations about the structure of the terrain~\cite{williams2001towards} and satellite imagery~\cite{garau2009path} to estimate the effects of currents. Genetic algorithms \cite{alvarez2004evolutionary} and mixed integer linear programming approaches \cite{yilmaz2008path} have also been used to support the computation of paths in dynamic underwater environments.

The work of Hernandez \etal\cite{hernandez2019online} provided an online sampling based framework for an AUV in 3D, accounting also for the dynamics of the system. The proposed framework although was shown to be capable to work in real-time, it needed close to half a minute for producing solutions with clearance guarantees. Moreover, during the online experiments the replanning was close to 1Hz but the robot was constrained to a constant depth reducing the planning space to 2D. When accounting for currents, other studies~\cite{lee2017energy,vidal2019online} utilize sampling\hyp based techniques and a complete dynamic model of the AUV, but are only shown to work in uncluttered environments with few obstacles.

Another challenge in underwater path planning is to generate paths rapidly enough to be able to compute and execute them online. Green and Kelly demonstrated a branching\hyp based method for quickly generating safe paths in a 2D environment  \cite{green2007toward} which has since been the basis of several optimizations  \cite{knepper2009empirical, knepper2012real, pivtoraiko2009differentially, branicky2008path}. Path planning has also been optimized by reducing many candidate paths to equivalence classes \cite{knepper2012toward}.

Though optimal sampling\hyp based techniques~\cite{karaman2010incremental, gammell2014informed, janson2015fast, gammell2015batch} provide near-optimal solutions and have improved over time, they, in general, require more computational resources, more time, and often an exhaustive search of the configuration space. Some studies on online underwater navigation with sampling-based techniques quickly generate safe paths, however, they are limited to 2D motions~\cite{hernandez2016planning} or require additional assumptions regarding the vertical relief~\cite{vidal2018online} without exploiting the full potential of available 3D motions. 

In some applications, it is necessary for AUVs to navigate in an environment without global knowledge of the environment. In such cases, obstacles are observed, often by stereo vision as has been done on aerial vehicles \cite{zhu20163d}. Exploration of an unknown environment by aerial vehicles has been performed using a 3D occupancy grid using probabilistic roadmaps (PRM) and the D* Lite algorithm for planning \cite{hrabar20083d}. Although underwater and aerial domains provide different challenges, both require path planning in 3D. For an AUV such as Aqua2, whose movements do not correlate exactly with control inputs, planning becomes even more difficult. Other AUVs have also been used for path planning, such as RAIS~\cite{antonelli2001real} and DeepC~\cite{eichhorn2005reactive}. Another AUV, REMUS~\cite{fodrea2003obstacle}, used obstacle avoidance specifically for exploration of shallow waters.

The Aqua2 AUVs have a variety of swimming gaits in order to enable tasks such as  swimming in a straight line, on the side, in a corkscrew motion, or performing a barrel roll~\cite{meger2015learning}. Visual tags were used to enable the AUV to navigate over structures~\cite{6942867}. Furthermore, the robot learned a reactive controller for navigating over the coral reef while maintaining a safe distance~\cite{Manderson2018_oceans,Manderson2018}.

\section{Proposed Approach}
\begin{figure}[t]
\centering 
{\includegraphics[width=0.4\textwidth, trim=1cm 7.1cm 1.75cm 1.2cm, clip=true]{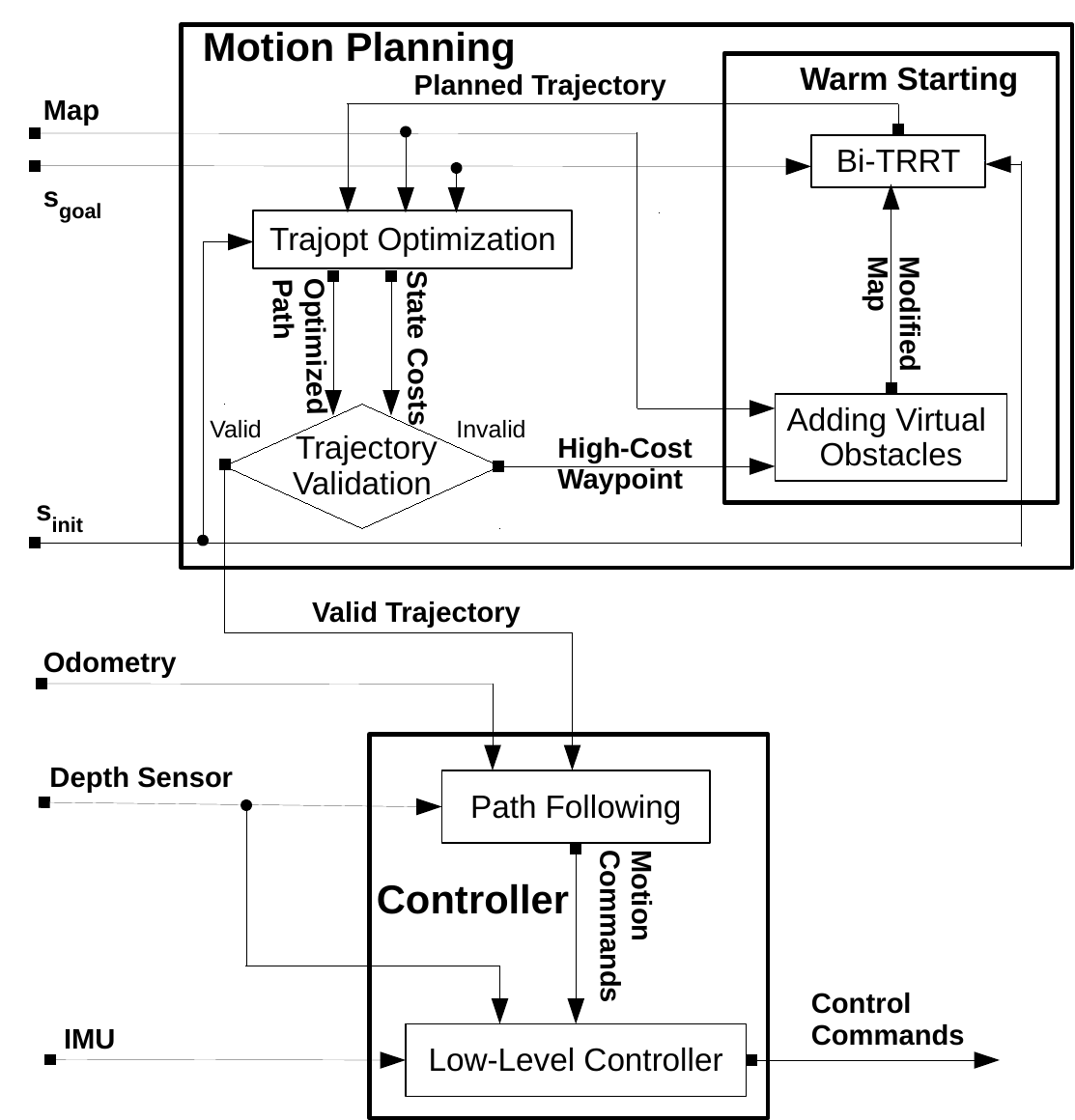}}
\caption{System architecture.}
\label{fig:pipeline}
\end{figure}

The objective of this paper is to develop algorithms that enable the Aqua2 robot to navigate reliably and safely through a dense field of obstacles, given start and goal poses $s_{\rm init}$ and $s_{\rm goal}$. Figure~\ref{fig:pipeline} presents an overview of the proposed system.

\subsection{System Overview}
The target system is the Aqua2 amphibious platform~\cite{dudek2007aqua}, pictured in \fig{fig:beauty}.
In its aquatic configuration, Aqua2 uses the motion from six flippers, each independently actuated by an electric motor, to swim.  Aqua2 has 6 degrees of freedom, of which five are controllable: two directions of translation (forward/backward and upward/downward), along with roll, pitch and yaw.

The robot's primary sensing modality is vision.  It is equipped with three iDS USB 3.0 UEye cameras: two facing forward and  one in the back. The front-facing cameras are used for navigation and data collection.
In addition to these cameras, Aqua2 also has an IMU and a pressure sensor which are used for controlling the motions and can be utilized for visual\hyp inertial state estimation~\cite{Huang2019ICRA,shkurti2011state,mourikis2007multi,RahmanICRA2018}. The fields of view of the cameras are 120 degrees (horizontal) and 90 degrees (vertical) tilted downward by 40 degrees from the horizontal plane.

\subsection{Trajectory Planning}
Motion planning in this context must balance several competing constraints, including the need for efficiency, the possible need to replan, and the limited computational power available on the robot (particularly when one considers the other essential tasks of perception, mapping, etc.).
In addition, because of its complex ---yet not readily modeled--- dynamics and kinematics combined with the unpredictable nature of maritime currents, the system is quite susceptible to disturbances. Thus, the planner should provide solutions that satisfy some minimum clearance to avoid collisions.
%


To satisfy this challenging trade-off, the proposed system utilizes an optimization-based planning approach. 
Specifically, the implementation uses Trajopt~\cite{schulman2013finding}, which has been proven as a very robust method for manipulators and mobile manipulators in 2D. It is computationally efficient and takes into account not only the states but the complete path between them using the transition between states, contrary to alternatives such as CHOMP~\cite{ratliff2009chomp} and STOMP~\cite{kalakrishnan2011stomp}. To the best of authors' knowledge, Trajopt has not yet been used for 3D motion planning for mobile robots, nor in the underwater domain, in the past.

The main idea behind Trajopt is to represent the path $S$ from $s_{\rm init}$ to $s_{\rm goal}$ as an ordered list of waypoints, each of which is a pose for the robot.  Starting from an initial set of waypoints ---generally based on the linear interpolation between $s_{\rm init}$ and $s_{\rm goal}$--- Trajopt forms a convex optimization program, in which each degree of freedom of each waypoint is a variable, the obstacles are encoded as constraints, and the objective is to minimize a weighted form of the path length. An important advantage is that, because of this general form, one can insert additional content-specific constraints, such as a maximum depth. 

\subsubsection{Input Methods}
\begin{figure}[t]
\begin{center}
\leavevmode
\begin{tabular}{ccc}
 \subfigure[]{\fbox{\includegraphics[height=2.2in, trim=20cm 2cm 20cm 2cm, clip=true]{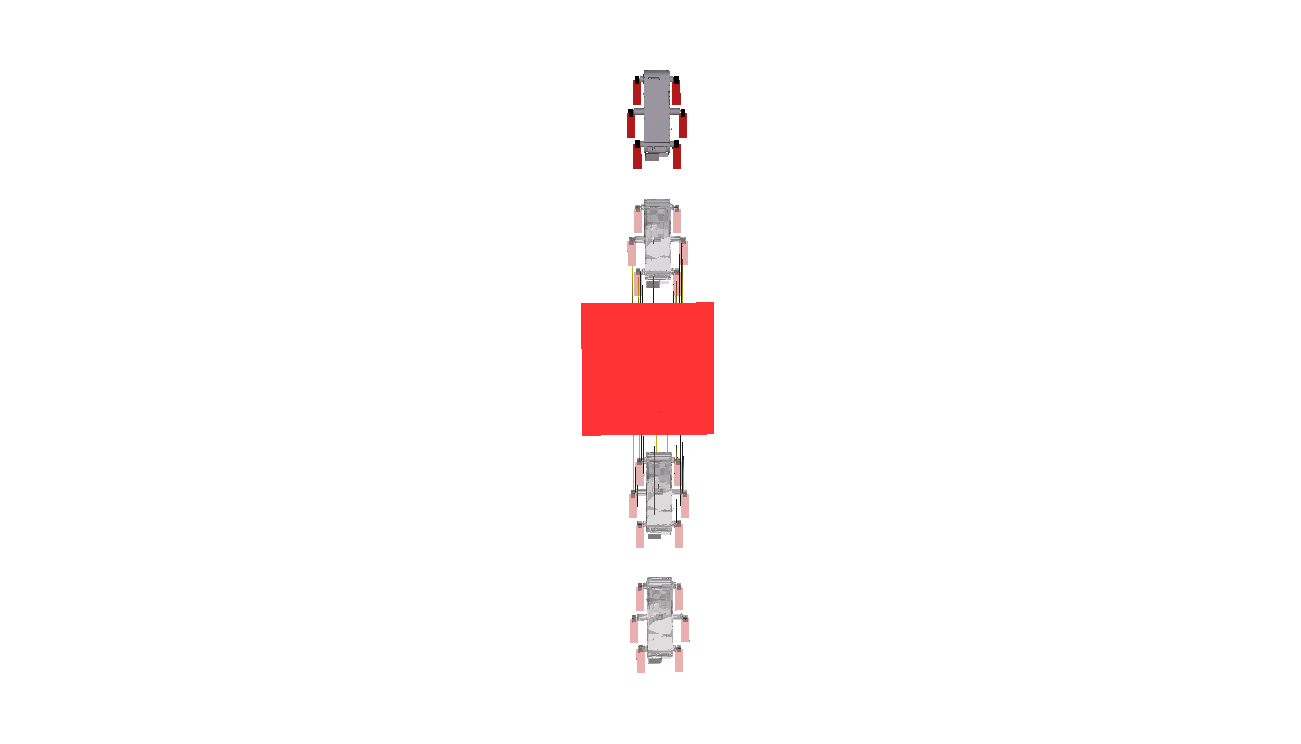}}}&
  \subfigure[]{\fbox{\includegraphics[height=2.2in, trim=20cm 2cm 19cm 2cm, clip=true]{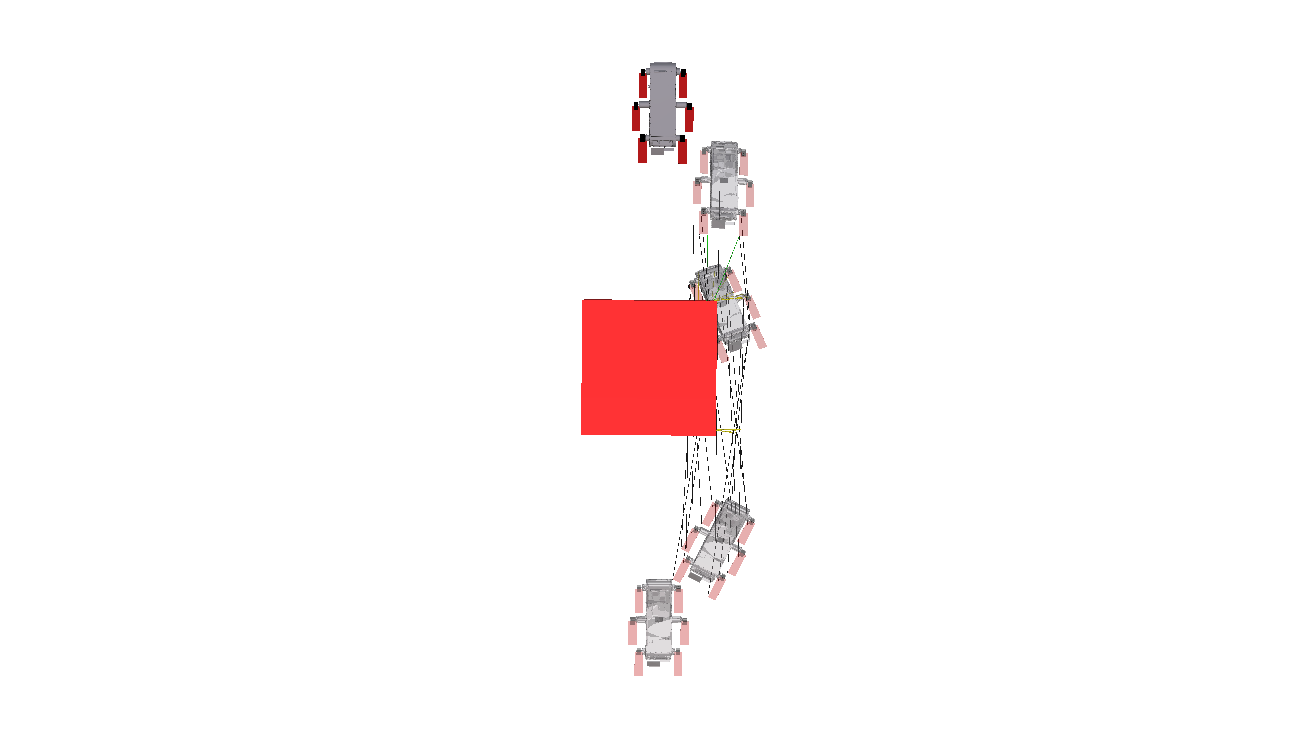}}}&
\subfigure[]{\fbox{\includegraphics[height=2.2in, trim=20cm 2cm 15cm 2cm, clip=true]{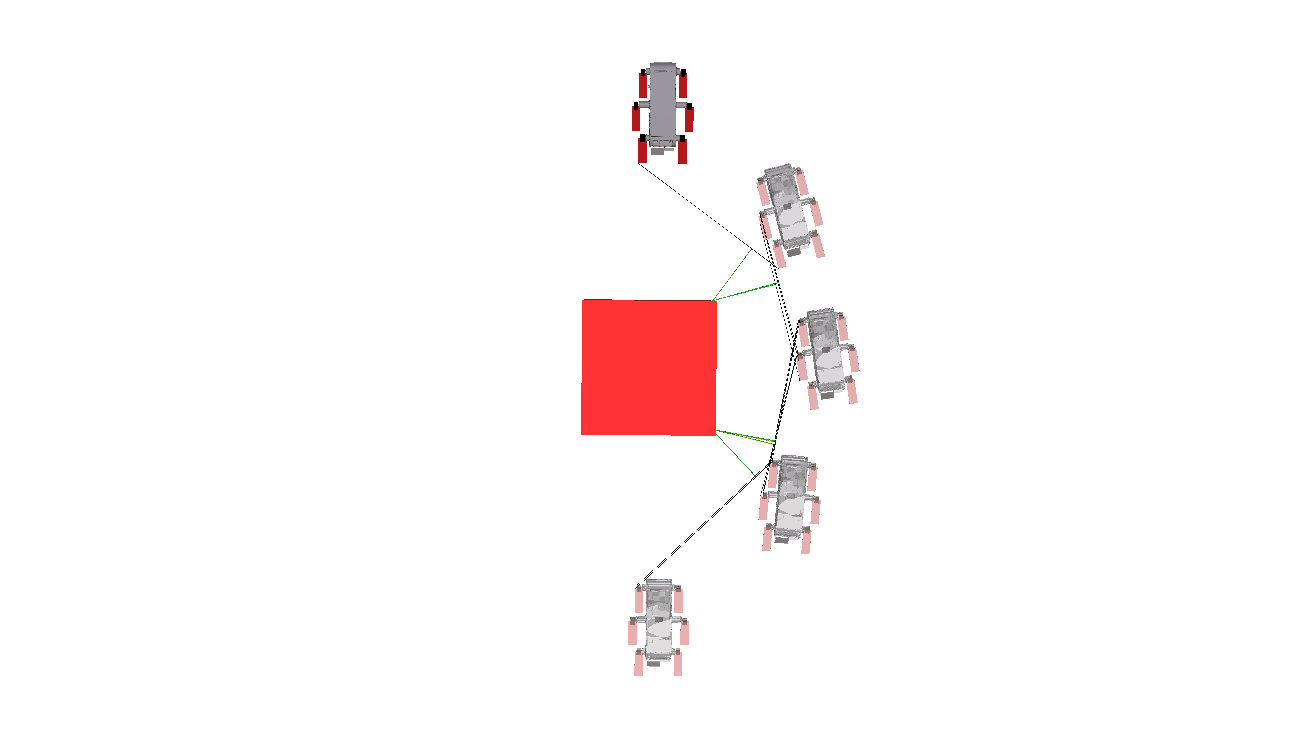}}}
\end{tabular}
 \vspace{-0.1in}  \caption{Path optimization with Trajopt~\cite{schulman2013finding} in three different stages: (a) The initial path is generated by simple interpolation from $s_{\rm init}$ to $s_{\rm goal}$ with possibly some states in collision.  (b) An intermediate stage during optimization.  (c) The final trajectory, shown with the distances to the swept-out volumes.}
   \label{fig:trajopt}
   \end{center}
 \end{figure}

Trajopt optimizes the distance between the swept-out volumes of the robot's trajectory and the obstacles; as shown in Figure~\ref{fig:trajopt}. For efficiency, Trajopt expects the map to be stored in a form where the normal vectors between the robot's body and the obstacles can be extracted rapidly.
One geometric method for presenting the obstacles to Trajopt, suitable in cases where the environment is well-known, consists of specifying the obstacle shapes and locations as instances of a set of built-in primitives, typically as rectangular boxes.

We also implemented a version that utilizes a point cloud input representation, which can be produced directly from raw sensor data.
First, the point cloud is provided as input to the fast surface reconstruction algorithm of Zoltan et~al.~\cite{Marton09ICRA}.
Then, the algorithm approximates the produced mesh via a collection of convex polytopes which can be directly processed by the Trajopt optimizer.
%
%

\subsubsection{Constraints}

The objective function is parameterized by coefficients for the path length and the obstacle avoidance and by a distance parameter $D_{\rm min}$, measuring the maximum distance from the obstacles where the cost will be applied.  These parameters required tuning for the underwater environment and the AUV used; Section~\ref{sec:exp} describes the specific values utilized in our experiments.

Our system employs an additional term in the cost to ensure that the robot will not reach the surface, and will  remain entirely underwater. The cost function $c_z$ was applied on the $z$ coordinate of all the states $s_i \in S$, where $s_i = [x_i,y_i,z_i, q^x_i,q^y_i,q^z_i,q^w_i]$, and defined as follows:
\begin{equation}
\label{c_z}
c_z(z_i) = 
\begin{cases}   
    z_i+\epsilon & \text{if $z_i> -D_{\rm min}$}\\
    0  & \text{otherwise}
\end{cases}
\end{equation}
where $\epsilon$ $> 0$. 
Assuming that on the surface $z=0$, the condition in Equation~\ref{c_z} penalizes every state above $-D_{\rm min}$ ensuring that the robot will remain underwater, accounting also for potential inaccuracies in control.

\subsection{Overcoming Local Minima}
A problem that many optimization-based motion planners, including Trajopt, face is the possibility that the optimization may converge to a local minimum. 
Though generally rare in the implemented system, this situation can present a safety hazard for the robot, because the completed path may not necessarily maintain safe distance from the obstacles. Local minima can be present, for example, either (a) when the path is passing through an obstacle and there is no free space in a $D_{\rm min}$ radius, or (b) when the path is passing through a narrow passage and the desired clearance from the obstacles cannot be maintained.
Typically, such issues are resolved with warm-starting of the optimization process: a fast motion planner is used to generate beforehand a set of valid paths and these paths are used consecutively to initialize the optimization until a valid solution is found. An example of such work~\cite{li2016birrtopt} for Trajopt uses the BiT-RRT~\cite{4650993} planner.

We propose the following novel \textit{iterative} warm\hyp starting approach, using BiT-RRT~\cite{4650993}, by inserting virtual obstacles into the planning process. The optimization step works with the motion planner in the following manner:
\begin{enumerate}
    \item The optimization result is checked and in case of failure, the waypoint with the highest cost is selected.
    \item The map used by the motion planner is altered by adding a virtual obstacle of size $D_{\rm min}$ in the position of that waypoint. 
    \item The planner plans a path avoiding the new high cost area of local minimum with a smaller than $D_{\rm min}$ distance from obstacles during collision checks.
    \item The new path is used to initialize the path optimization process and the procedure continues if needed. 
\end{enumerate}

\noindent This process forces the planner to identify an alternative path that avoids the most problematic portion, in terms of the cost function, of the previous path.

\subsection{Trajectory Following}

\subsubsection{Localization}
 The problem of underwater localization has  proven to be extremely challenging~\cite{JoshiIROS2019,QuattriniLiISERVO2016} due to the lighting variations, hazing, and color loss~\cite{SkaffBMVC2008}.  We tested two solution strategies for this problem.
\paragraph*{Primitive Estimator} A primitive estimator has been employed using the depth sensor, the attitude estimation from the IMU and the expected forward speed of the AUV (based on the swimming pattern utilized). Though subject to drift over long distances, this estimator has guided the Aqua2 vehicles for a variety of basic maneuvers and swimming patterns. During operations with a known map, this primitive estimator is utilized to track the planned trajectory. 

\paragraph*{SVIn} During operations in an unknown environment, more accurate localization may be useful, in addition to the required ability to detect nearby obstacles.  Visual Inertial state estimation, introduced by Rahman \etal \cite{RahmanICRA2018} and later extended to utilize depth measurements and loop\hyp closures~\cite{RahmanIROS2019a}, is used for simultaneously localizing and mapping nearby obstacles. The resulting point\hyp cloud produced by SVIn enables the AUV to navigate safely around obstacles.  

\subsubsection{Waypoint seeking}
The optimization stage of the proposed pipeline produces in a timely manner a path $p$, as an ordered set of consecutive goal states that should be sequentially achieved by the robot. 
A linear PD controller proposed by Meger et al.~\cite{6942867} is utilized, which employs the IMU and the depth sensor data. This closed-loop controller accepts commands in the form:
\begin{equation}
    \label{tuple}
    com = \left\langle v, h, d, o \right\rangle,
\end{equation}
\noindent in which $v$ is the desired forward linear velocity, $h$ is the desired heave (that is, upward or downward linear velocity), $d$ is the desired depth to reach and $o$ the desired orientation for the robot to move in the world frame. The current framework, for simplicity considers only purely forward motion setting $h=0$.  

The PD controller controls the desired depth by linear interpolation.  Assuming that the current state of the robot is $s_c$ and the current $i$ goal position is $p_i$ in the world frame, to guarantee smooth transitions that are bounded inside the calculated safe swept-out volumes of the optimization stage, the desired depth $d$ is calculated as
\begin{equation}
    \label{depth}
    d = z_{p_{i-1}} + \left(1-\frac{z_{s_c}}{z_{p_{i-1}}}\right)\left(z_{p_{i}}-z_{p_{i-1}}\right),
\end{equation}
where $z_{p_{i-1}}$ is the depth of the previous achieved goal (base case $z_{p_{i-1}} = s_{\rm init}$),  $z_{p_{i}}$ the depth of the current goal, and $z_{s_c}$ is the current measured depth. The idea is to ensure the linear change of  depth from one position to the other and ensure linear transitions similar to the ones assumed by Trajopt.

Regarding the desired orientation, the pitch is adjusted automatically from the desired depth and the roll does not affect the direction of the motion, thus only the computation of the desired yaw $o_{\rm yaw}$ is needed, with respect to the translation error $e_t = p_{i}-s_c$. Given the position of the robot, the yaw changes in such a way that the AUV will always move towards facing the goal.

Lastly, assuming the possible deviation is bounded by $D_{\rm min}$ for a given speed and that the optimization was successful, the AUV should safely navigate from one goal to the next one. As a result of the above, given a threshold of $D_{\rm reached}$, the goal is declared reached, if the error $e_t$ is less than $D_{\rm reached}$ and a local minimum is detected, since otherwise the robot will be sliding away due to disturbances.

\begin{figure}[ht]
\centering
\leavevmode
\begin{tabular}{lr}
\subfigure[]{\includegraphics[height=0.83in]{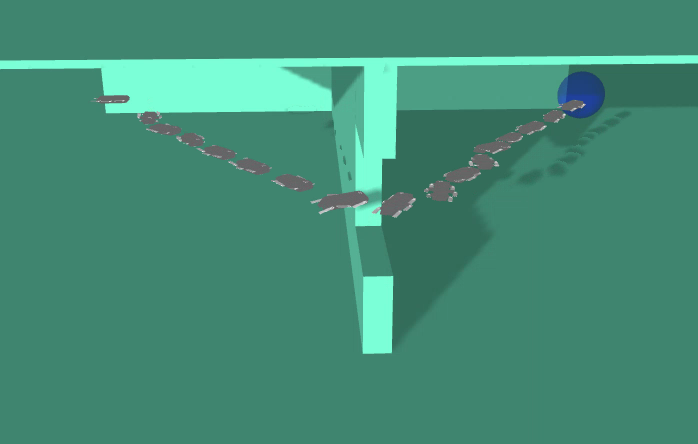}\label{fig:simRoom}}&
\subfigure[]{\includegraphics[height=0.83in]{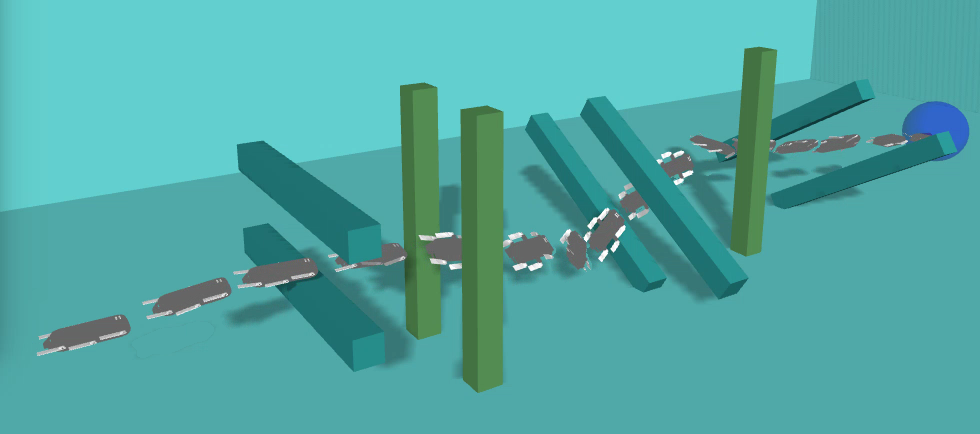}\label{fig:simComplex}}
\end{tabular}
\vspace{-0.1in}   \caption{Simulated trajectories executed by the robot in Gazebo. \subref{fig:simRoom} An environment with an narrow opening (the ceiling is not shown); \subref{fig:simComplex} A cluttered environment with multiple pipes. }
   \label{fig:SimulatedOffline}
 \end{figure}
 \begin{figure*}[t]
\centering
\leavevmode
\begin{tabular}{cccc}
 \subfigure[]{\includegraphics[height=1.5in]{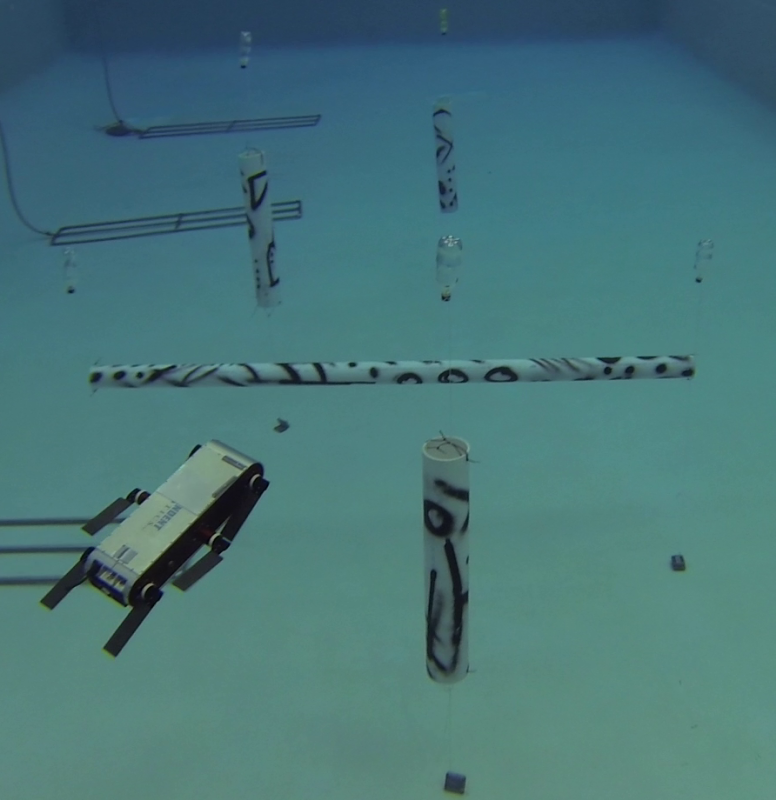}\label{fig:pool1i}}&
  \subfigure[]{\includegraphics[height=1.5in]{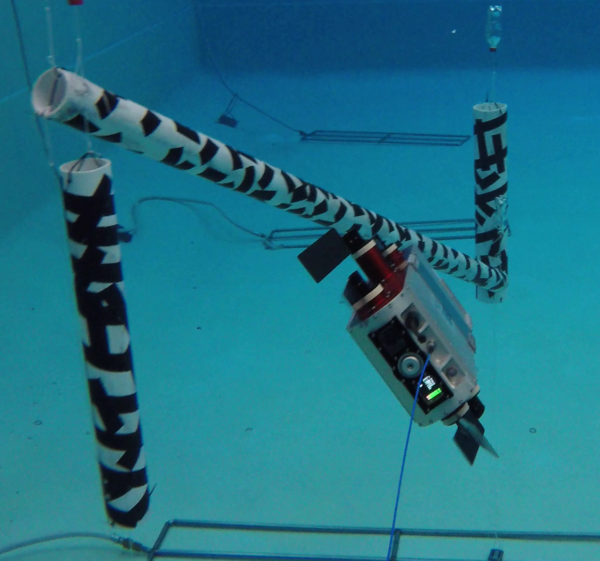}\label{fig:pool2i}}&
\subfigure[]{\includegraphics[height=1.5in]{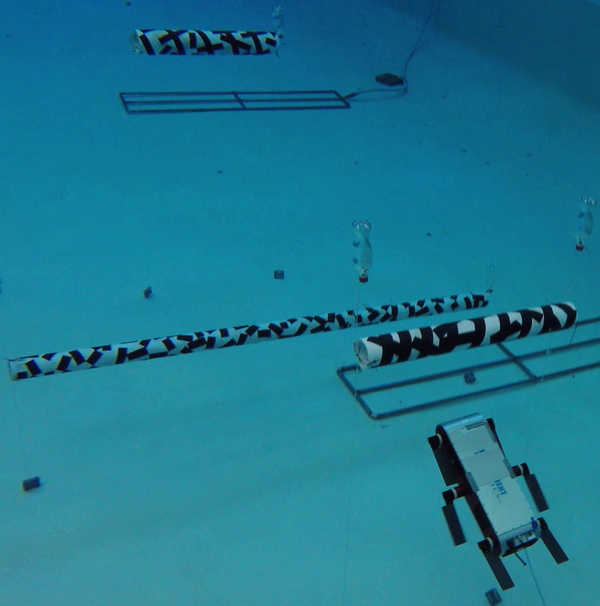}\label{fig:pool3i}}&
\subfigure[]{\includegraphics[height=1.5in]{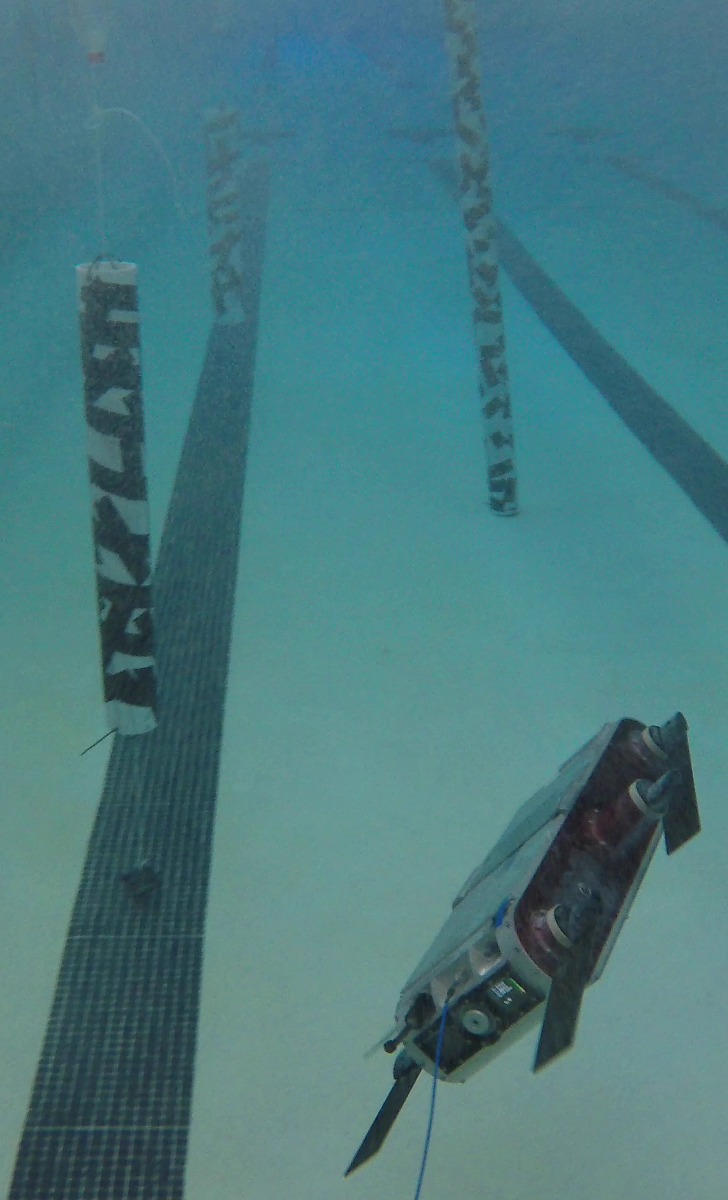}\label{fig:pool4i}}\\
 \subfigure[]{\includegraphics[height=0.5in]{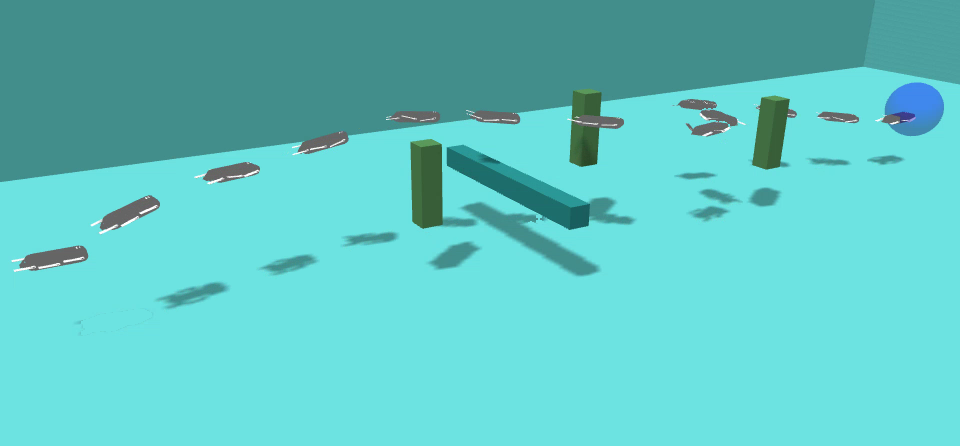}\label{fig:poolMap1}}&
  \subfigure[]{\includegraphics[height=0.5in]{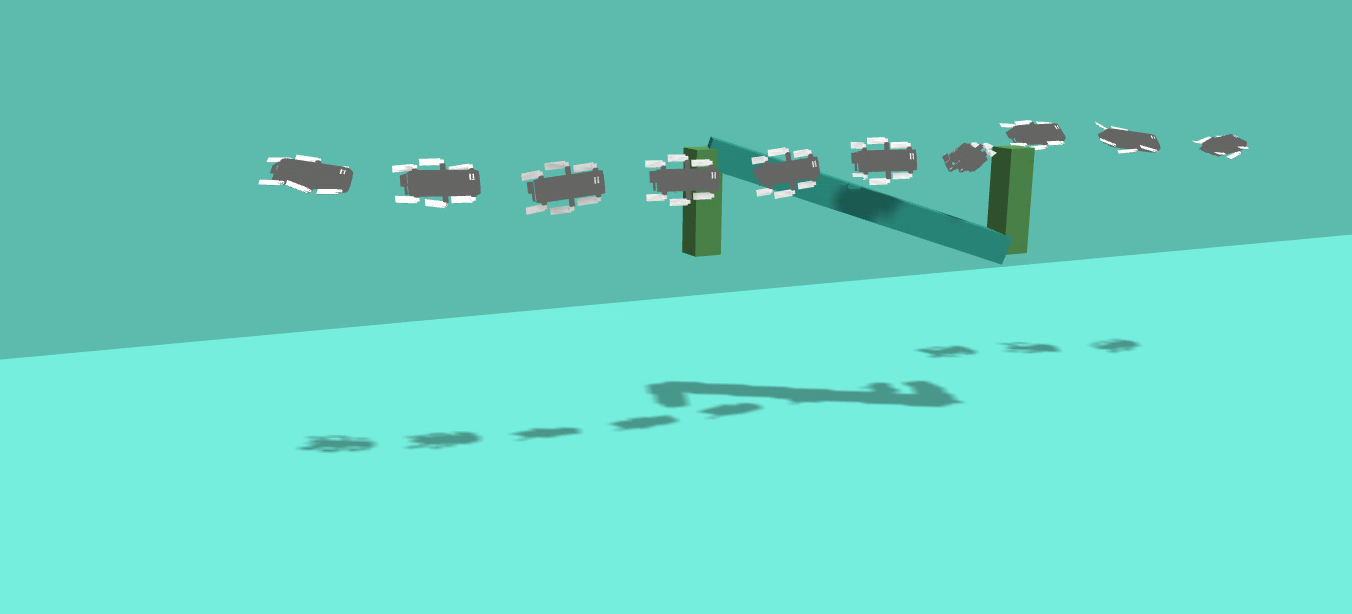}\label{fig:poolMap2}}&
\subfigure[]{\includegraphics[height=0.5in]{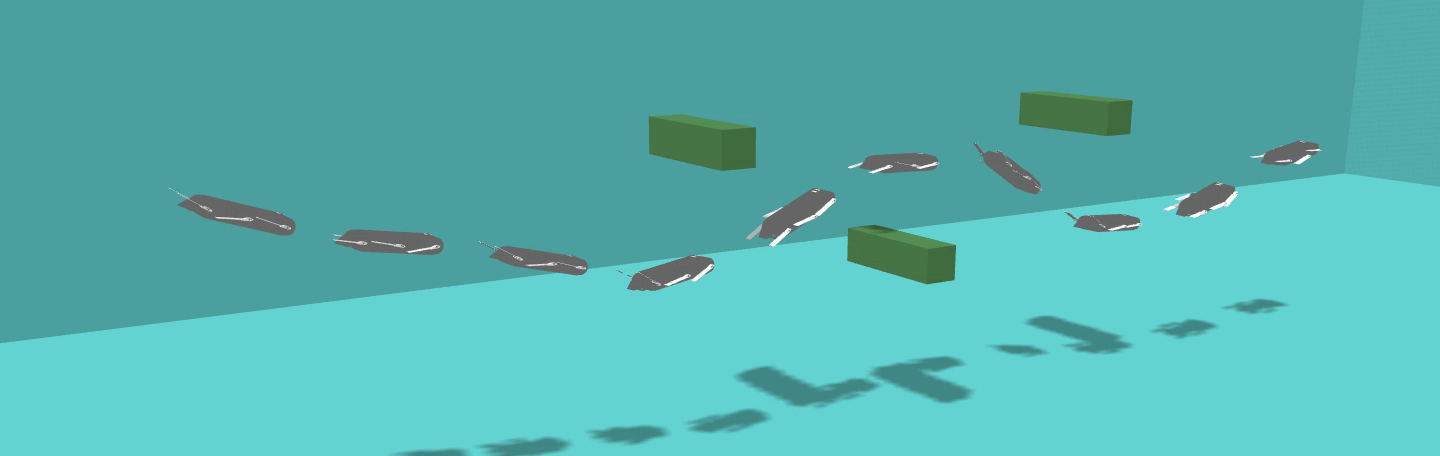}\label{fig:poolMap3}}&
\subfigure[]{\includegraphics[height=0.5in]{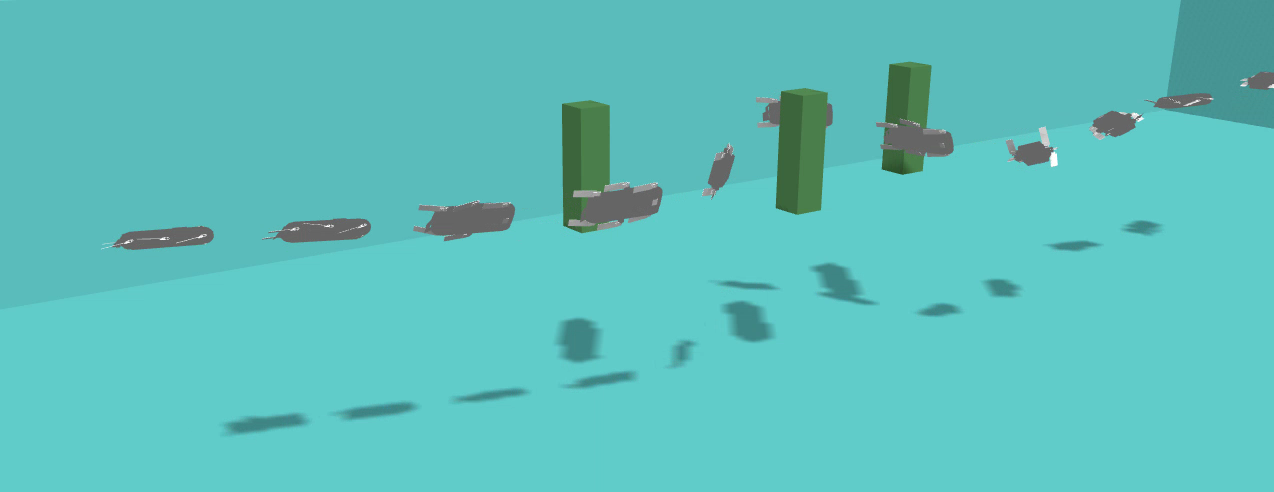}\label{fig:poolMap4}}\end{tabular}
   \caption{Top-row shows representative photos from the real deployment in the pool for four different environments. The first three environments where in the deep diving pool and the fourth in the shallower swimming pool: \subref{fig:pool1i} AUV passes between the vertical obstacles and over the horizontal obstacle; \subref{fig:pool2i} the planner generates a $45^\circ$ roll for the AUV while passing between the two vertical obstacles and over the diagonal one; \subref{fig:pool3i} AUV goes under the first, over the second, and under the third horizontal obstacle, and the planner guides a flat roll; \subref{fig:pool4i} AUV again has a $45^\circ$ roll passing around the three vertical obstacles.  The second row shows a view of the executed path in Gazebo for demonstration purposes.}
   \label{fig:RealOffline}
 \end{figure*}

\section{Experimental Results}
\label{sec:exp}
Extensive experiments were performed using the Gazebo simulator, and in numerous deployments of the Aqua2 AUV at two pools in our university, a shallow swimming pool of dimensions $50m\times25m\times2m$ and a deep diving pool of dimensions $25m\times15m\times4m$. The main objective of the various experiments was to demonstrate the reliable navigation functionality of Aqua2 using the proposed framework. In all of the experiments the AUV had a constant speed of $0.4 m/s$ --- the expected operational speed --- and a bounded motion with a minimum obstacle avoidance distance of $D_{\rm min} = 0.6\si{m}$. Obstacle avoidance and path length coefficients were adjusted to relatively high values $200$ and $100$, respectively, favoring safety over path length optimality, while the number of waypoints was determined either by the distance from the current position to the goal --- placing a state every $1.5 m$ --- or by the number of states provided by the Trajopt planner. The experiments tested planning on a known map, with a focus on efficient trajectories, and online, using the camera for obstacle avoidance and frequent replanning.

\subsection{Known Map --- Offline Planning}
During planning with a known map, the input map was a set of geometric primitives (planes, boxes, cylinders, etc.) and the iterative warm\hyp starting process was used. All plans were produced  in less than half a second.

\subsubsection{Simulated Environment}
Two different environments are presented here highlighting different challenges for Trajopt. In the first environment, called the Window environment, two rooms, separated by a wall with an opening (window) between them and open to one side demonstrates path optimality; see \fig{fig:simRoom}. The original path results in a local minimum. After the {iterative warm\hyp starting} method is used the solution through the window is found. Furthermore, for this experiment the AUV has to keep a horizontal pose during motions that causes larger drifting from inertia. For this purpose, in this single case we increased $D_{\rm min}$ to $1\si{m}$.

The second environment, called the Cluttered environment, focuses more on the capability of our method, inherited from Trajopt, to minimize oscillation while passing through a sequence of obstacles that need accurate motions with fast orientation changes; see \fig{fig:simComplex}. In this environment not only was the roll adjusted during optimization adapting to the motion, but also the Aqua2  is guided to pass in a parallel motion  between each pair of pillars, thus maximizing the distance from both of them, for increased safety. 

\begin{figure}[ht]
\leavevmode
\begin{center}
\begin{tabular}{c}
\subfigure[]{\includegraphics[width=0.47\textwidth, trim=10.5cm 0.0cm 10.5cm 1.0cm, clip=true]{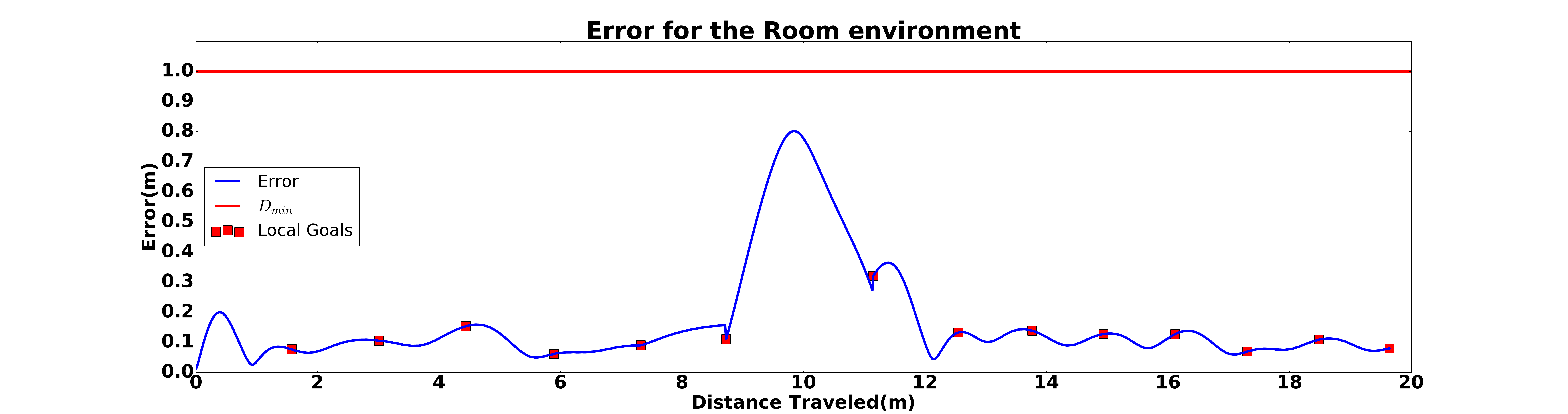}\label{fig:error_room}} \\
\subfigure[]{\includegraphics[width=0.47\textwidth, trim=10.5cm 0.0cm 10.5cm 1.0cm, clip=true]{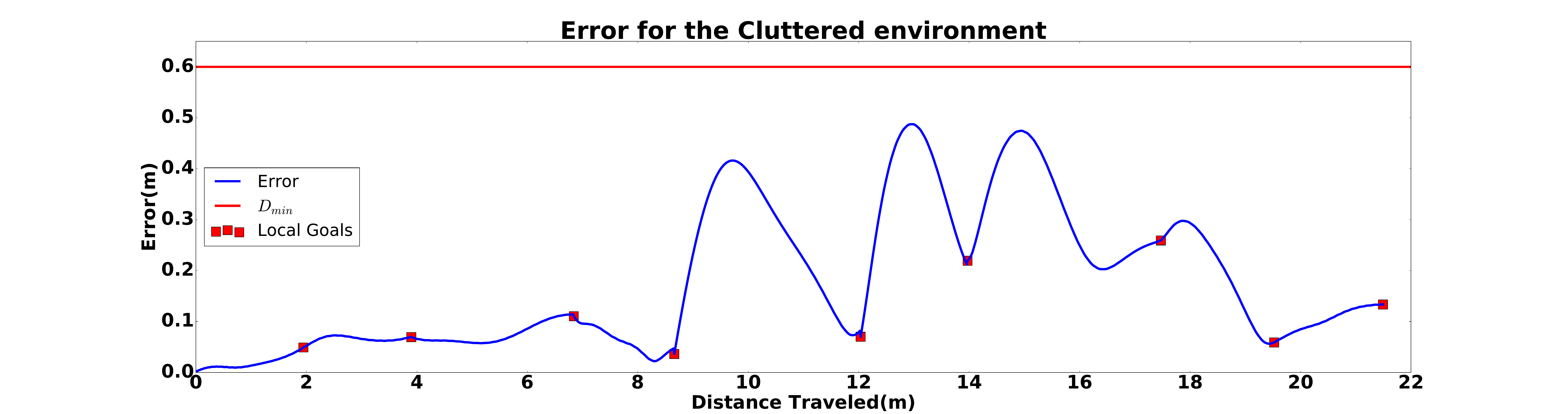}\label{fig:error_clut}}
\end{tabular}
\vspace{-0.1in}   \caption{Error diagrams for the Room~\subref{fig:error_room} and the Cluttered~\subref{fig:error_clut} environment, as measured from the simulation. The red squares indicate the local goals achieved, and the red line marks $D_{\rm min}$, where errors larger than that should be considered unsafe.}
  \label{fig:Sim_error}
  \end{center}
 \end{figure}

Safe navigation for the Aqua2 is achieved only if during the motion the trajectory following method does not violate the assumed clearance $D_{\rm min}$ by the planning process. For these challenging scenarios, where the robot changes orientations fast, the oscillations are shown in Figure~\ref{fig:Sim_error}. The error at time $t$, $e_t$, is calculated as the Euclidean distance of the measured simulation odometry $s_c$, to the line formed by the previous $p_{i-1}$ and the current local goal $p_i$:

\begin{equation}
    e_t = \frac{|(s_c-p_{i-1})\times(s_c-p_{i})|}{|p_{i}-p_{i-1}|}
\end{equation}

\begin{figure}[h]
    \centering
    \includegraphics[width=0.95\columnwidth]{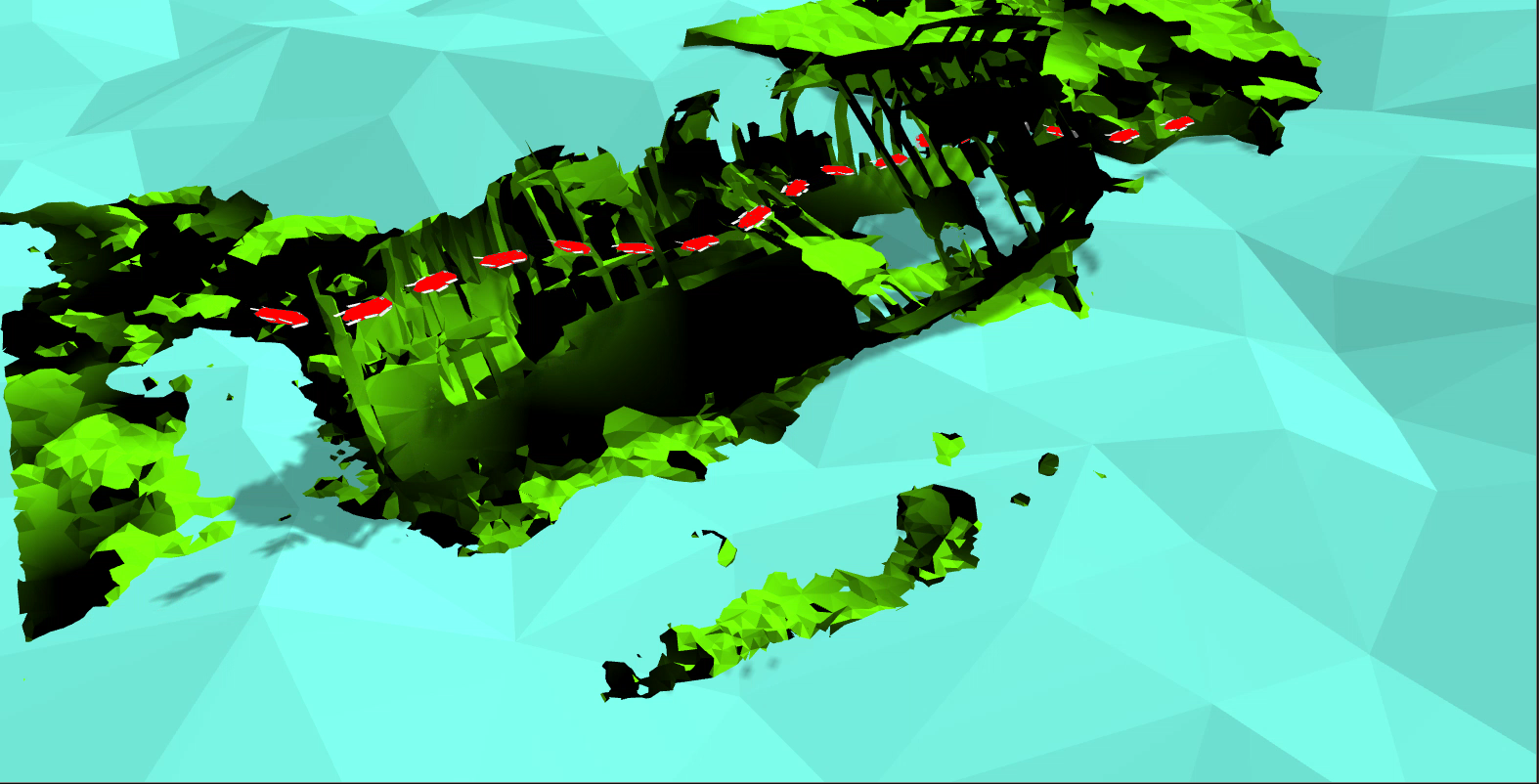}
    \caption{Navigating inside a shipwreck model in simulation. Model of ``Shipwreck, Hooe Lake, Plymouth'' from Sketchfab.}
    \label{fig:SimShipwreck}
\end{figure}

 \begin{figure*}[ht]
\centering
\begin{tabular}{lcr}
 \subfigure[]{\includegraphics[height=1.40in]{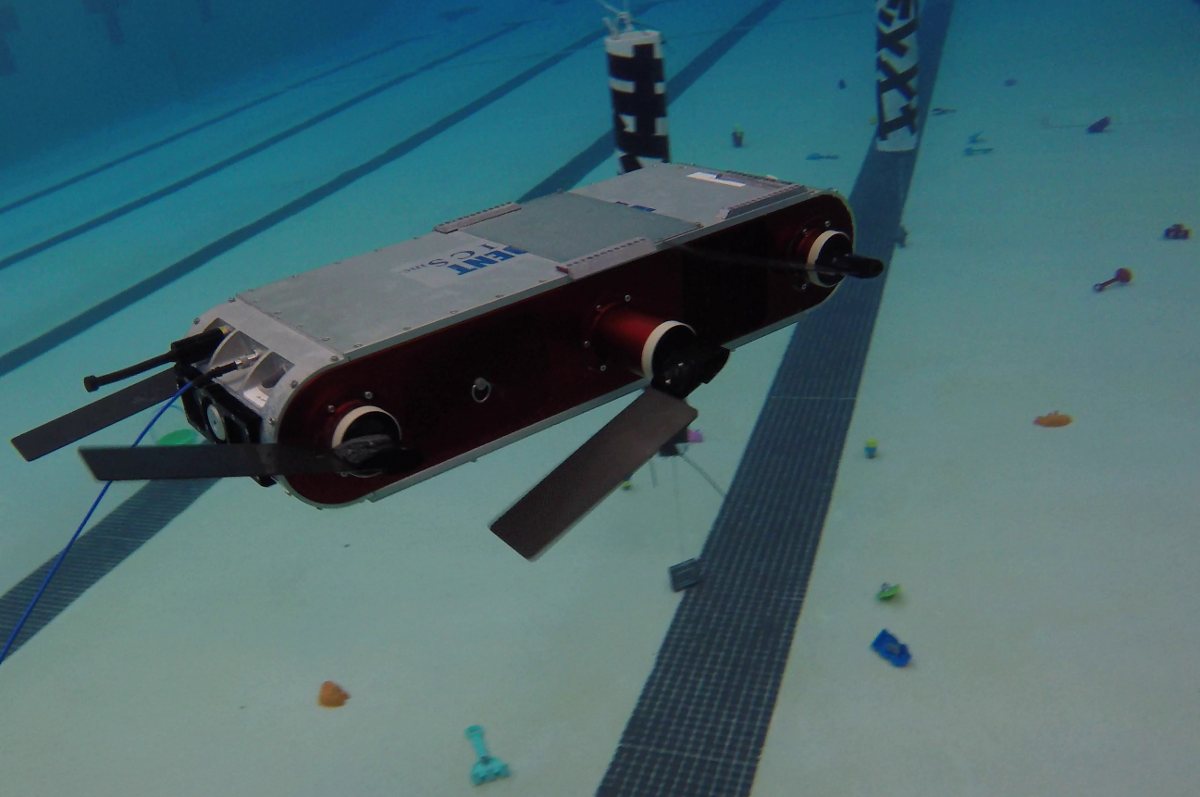}\label{fig:OnlinePoolImg1}}&
 \subfigure[]{\includegraphics[height=1.40in]{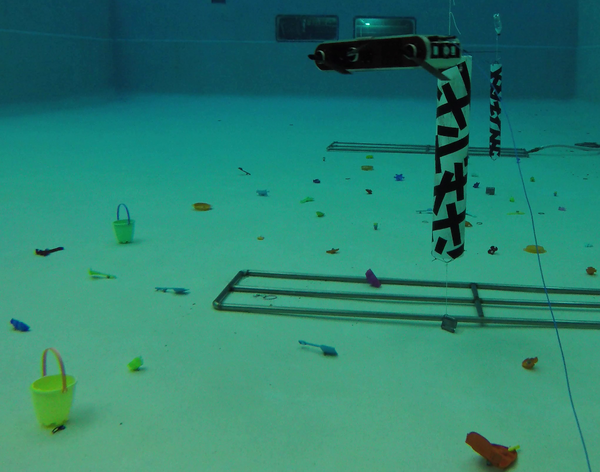}\label{fig:OnlinePoolImg2}}&
 \subfigure[]{\includegraphics[height=1.40in]{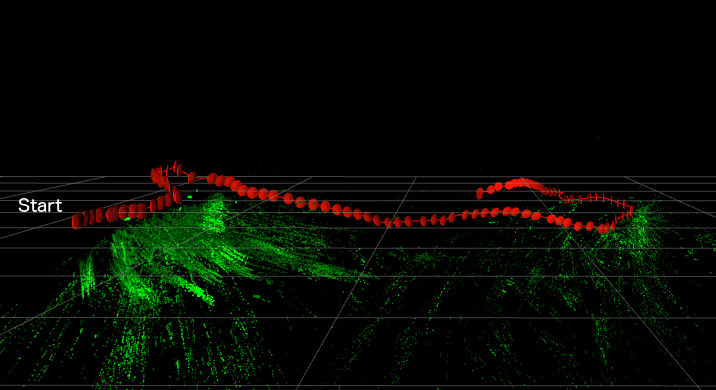}\label{fig:OnlinePoolRViz2}}
\end{tabular}
\vspace{-0.1in}    \caption{\subref{fig:OnlinePoolImg1},\subref{fig:OnlinePoolImg2} show representative photos from the deployments in the pool in an unknown environment. Please note the plastic toys spread at the bottom of the pool to produce detectable features in a featureless pool. \subref{fig:OnlinePoolImg1} Avoiding two obstacles in the shallow swimming pool; \subref{fig:OnlinePoolImg2} Avoiding two obstacles in the deep diving pool. \subref{fig:OnlinePoolRViz2} presents the online map produced by SVIn~\cite{RahmanICRA2018} as a screenshot of RViz for the environments of \subref{fig:OnlinePoolImg2}, the robot avoid the first cylinder, moves forward and then avoids the second, while using the features from the bottom of the pool to localize.}
   \label{fig:OnlinePool}
 \end{figure*}

\subsubsection{Real}
Four different environments were used at the two pools in our university, demonstrating operations with the Aqua2 AUV; see \fig{fig:RealOffline}. The placement of the obstacles followed the created map used as input to the augmented Trajopt planner. In the first environment, the Aqua2 was forced to maintain an attitude facing the floor, a situation that maximizes drift during yaw changes. The robot was able to successfully avoid all the obstacles and quickly self\hyp correct its orientation; see \fig{fig:pool1i} for a image of the AUV during operations and \fig{fig:poolMap1} for an overview of the trajectory. For the other three environments, the attitude of the AUV was optimized by the planner. In the second environment the robot avoided the two vertical pipes and passed over the diagonal one; see \fig{fig:RealOffline}(b),(f). Three horizontal pipes were used to force the AUV in continuous depth changes; see \fig{fig:RealOffline}(c),(g). Finally, three vertical pipes resulted again in obstacle avoidance with a roll at $45^\circ$; see \fig{fig:RealOffline}(d),(h).

\subsection{Sensor based Planning --- Online}
The deployment of the proposed framework online highlighted some computational challenges. First, the state estimation consumes a large fraction of the computing resources. Second, the most computationally expensive component of the motion planning pipeline is the convex decomposition step, using approximately $70-80\%$ of the total planning time. For efficiency, the convex decomposition parameters were adjusted to produce many small convex polyhedra, instead of a few large ones. A history of the observations was kept to account for the limited field of view of the AUV.  In all cases the replanning frequency was on average at 1Hz.

\subsubsection{Simulated Environment}
The model ``Shipwreck, Hooe Lake, Plymouth'' from Sketchfab\footnote{\url{http://sketchfab.com/}} was simplified and used in the Gazebo simulator. For computational efficiency, no texture was used in the simulator and the basic point\hyp cloud was acquired from the model using a resolution of $100\times75$ points. The AUV was guided through the inside of the shipwreck, while avoiding the observed obstacles; see \fig{fig:SimShipwreck}. The complex environment presented in \fig{fig:simComplex} was used for online navigation resulting in similar results.

\subsubsection{Real}
Additional real pool experiments were conducted using state estimation from  the robust underwater SLAM package SVIN~\cite{RahmanIROS2019a} with additional sand\hyp toys weighted and placed on the floor to improve the  odometry estimation together with obstacles to test obstacle avoidance; see \fig{fig:OnlinePool}. The same configuration was used with the online simulation framework with the difference that the AUV was constrained at constant depth to maintain tracking using the features on the floor. 
The two environments, at the shallow and deep pool, used two vertical obstacles resulting in similar obstacle avoidance trajectories. \fig{fig:OnlinePoolRViz2} presents the overall recorded trajectory and features detected from SVIn~\cite{RahmanIROS2019a}. During one of the experiments the AUV's motion brought it towards a diver recording the experiment, who, to the diver's relief, was treated as another obstacle and was avoided.

\section{CONCLUSIONS}
\label{sec:conc}
This paper demonstrated novel capabilities for underwater navigation through cluttered spaces. The proposed framework was able to plan and successfully execute efficient trajectories, while avoiding obstacles and taking into account the kinematic constraints of the vehicle, in both known and unknown environments. Numerous simulations highlighted the abilities of this agile platform to navigate in narrow spaces. Experiments at the pool demonstrated the feasibility of the proposed method and highlighted challenge.

\begin{figure}[h]
\centering
\includegraphics[width=0.4\textwidth]{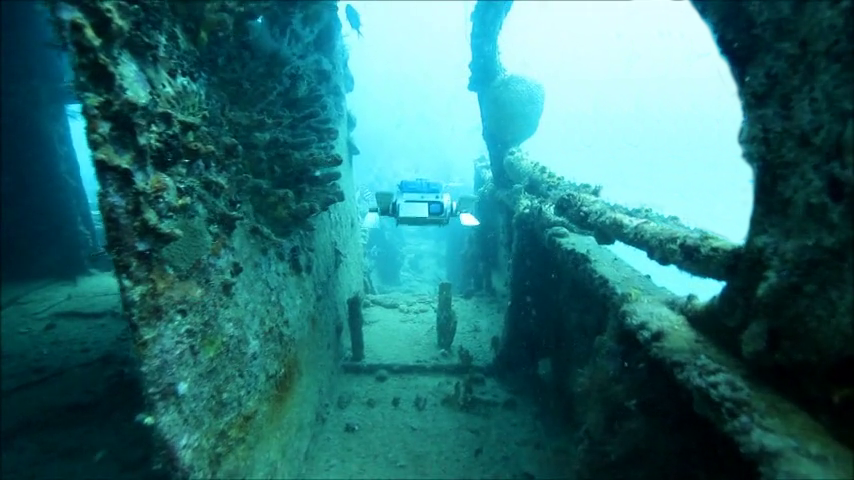}
\caption{Aqua2 AUV moving through a wreck.}
\label{fig:pamir}
\end{figure}

The presented results, combined with recent advancements in the robustness of underwater state estimation~\cite{RahmanIROS2019b} and with new advanced gaits for the Aqua2 vehicles~\cite{meger2015learning},
will enable future autonomous exploration of environments not previously accessible to robots, such as shipwrecks and caves.  For example, \fig{fig:pamir} presents an enclosed environment at the Pamir shipwreck near Barbados, for which no prior map exists. In the past, an AUV was deployed there to perform a simple straight line transect. Future work will test the proposed approach in deployments to more fully explore such environments.




\showtotalpagebudget





\bibliographystyle{IEEEtran}
\bibliography{./IEEEabrv,refs}


\end{document}